\documentclass[letterpaper]{article} 
\usepackage{aaai2026}  
\usepackage{times}  
\usepackage{helvet}  
\usepackage{courier}  
\usepackage[hyphens]{url}  
\usepackage{graphicx} 
\usepackage{natbib}  
\usepackage{caption} 
\usepackage{algorithm}
\usepackage{algorithmic}

\usepackage{amsmath}
\usepackage{amssymb}
\usepackage{multirow}

\usepackage{newfloat}
\usepackage{listings}
\DeclareCaptionStyle{ruled}{labelfont=normalfont,labelsep=colon,strut=off} 
\floatstyle{ruled}
\newfloat{listing}{tb}{lst}{}
\floatname{listing}{Listing}
\title{PaintFlow: A Unified Framework for Interactive Oil Paintings Editing and Generation}
\author{
    Zhangli Hu, Ye Chen, Jiajun Yao, Bingbing Ni\thanks{Corresponding author.}
}

\affiliations{
    School of Information Science and Electronic Engineering, Shanhai Jiao Tong University, China \\
    tension-2019@sjtu.edu.cn
}

\usepackage{bibentry}

\begin{document}

\maketitle

\begin{abstract}
Oil painting, as a high-level medium that blends human abstract thinking with artistic expression, poses substantial challenges for digital generation and editing due to its intricate brushstroke dynamics and stylized characteristics. Existing generation and editing techniques are often constrained by the distribution of training data and primarily focus on modifying real photographs. In this work, we introduce a unified multimodal framework for oil painting generation and editing. The proposed system allows users to incorporate reference images for precise semantic control, hand-drawn sketches for spatial structure alignment, and natural language prompts for high-level semantic guidance, while consistently maintaining a unified painting style across all outputs. Our method achieves interactive oil painting creation through three crucial technical advancements. First, we enhance the training stage with spatial alignment and semantic enhancement conditioning strategy, which map masks and sketches into spatial constraints, and encode contextual embedding from reference images and text into feature constraints, enabling object-level semantic alignment. Second, to overcome data scarcity, we propose a self-supervised style transfer pipeline based on Stroke-Based Rendering (SBR), which simulates the inpainting dynamics of oil painting restoration, converting real images into stylized oil paintings with preserved brushstroke textures to construct a large-scale paired training dataset. Finally, during inference, we integrate features using the AdaIN operator to ensure stylistic consistency. Extensive experiments demonstrate that our interactive system enables fine-grained editing while preserving the artistic qualities of oil paintings, achieving an unprecedented level of imagination realization in stylized oil paintings generation and editing.
\end{abstract}


\begin{figure}[t]
  \includegraphics[width=0.48\textwidth]{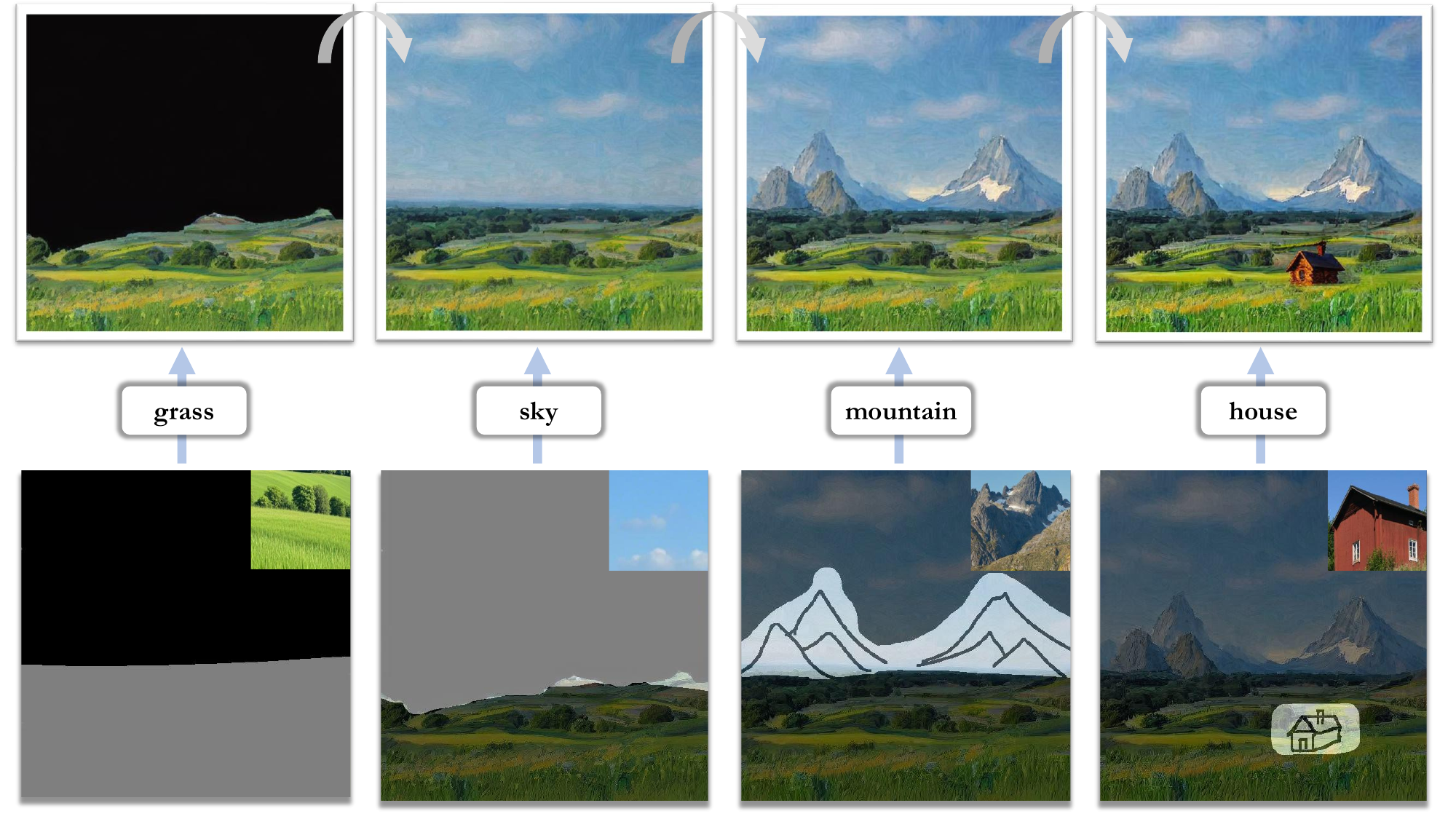}
  \centering
  \caption{PaintFlow is an interactive oil painting editing and generation system. It enables users to flexibly combine text, reference images, and sketches for fine-grained control. 
  Please zoom in to capture more details.}
  \label{fig:teaser}
\end{figure}

\section{Introduction}

Oil painting, as a vital carrier of traditional culture, holds irreplaceable historical and artistic value. However, its high level of professional expertise presents a barrier to the widespread adoption of traditional creation methods, limiting the people’s involvement in the creative process. In response to this issue, interactive painting technology has emerged, aiming to build a creative space that aligns with human cognitive needs and expands the dimensions of artistic expression~\cite{chen2015wetbrush, singh2022intelli, liu2024magicquill}. As a result, it has attracted wide attention across interdisciplinary fields such as design and education.

However, despite the rapid development of AIGC~\cite{rombach2022high, zhang2023adding, podell2023sdxl} and graphics technologies~\cite{chen2015wetbrush, wang2024physics}, interactive generation of oil painting styles still faces critical challenges: (1) The unique visual representation of oil painting emphasizes layered brushstroke textures rather than the precision of visual structure; (2) Creating a realistic oil painting requires the artist to dynamically adjust the content of the canvas through multimodal interactions such as sketches, text, and reference images, while maintaining the consistent collaboration of style across the entire canvas.
Current research focuses on two major paradigms: reconstruction and generation. The reconstruction paradigm~\cite{huang2019learning, tong2022im2oil, hu2023stroke, hu2024towards} relies on predefined brushstroke libraries to convert natural images into an oil painting style. While high-resolution works can be generated, they are limited to the visual reconstruction of existing content, do not support interactive user creation, and are constrained by fixed brushstroke libraries that cannot cover diverse artistic styles. The generation paradigmn~\cite{rombach2022high,zhang2023adding,ye2023ip,yang2023paint,liu2024magicquill, hertz2024style, liu2024stylecrafter}, based on models like Stable Diffusion~\cite{rombach2022high}, utilizes conditional injection for stylized output. While these methods support custom image editing, they are limited by the data distribution preferences of base models: they struggle to consistently generate structure-precise oil painting style images and lack high-quality multimodal annotated data, making it difficult to support core interactive methods such as sketch guidance and text descriptions.

To address these limitations, this paper proposes a novel unified interactive oil painting style image editing and generation framework. We select the Stable Diffusion model as the base model. Mimicking the creative process of human artists in oil painting, we have customized three interactive mechanisms for this model: text channel carries the thematic semantics of the creation, sketch channel locates the spatial topology structure, and reference image channel anchors the visual content entities. To support these three interactive mechanisms while ensuring that the generated content fully aligns with the user’s multimodal instructions and maintains a consistent oil painting style across the entire canvas, we have developed an oil painting style image diffusion model. The model design adheres to the principle of content-style decoupling. During the training phase, conditional alignment techniques map the mask/sketch to the channel dimension of the U-Net, while the reference image is encoded as contextual embedding through a cross-attention module, complemented by a semantic enhancement strategy to improve fine-grained feature fidelity. In the inference phase, a style alignment system is applied, aligning source domain style features through the AdaIN ~\cite{huang2017arbitrary}module, then merging the stylized context with the semantic vector output by the pretrained text encoder for cross-modal fusion. This solution systematically addresses the three challenges of multimodal coordination, style controllability, and structural consistency.

Comprehensive experiments demonstrate that PaintFlow shows significant advantages in both multimodal instruction alignment and painting style preservation, with performance metrics surpassing current mainstream methods. A groundbreaking aspect of our framework is its ability to achieve full-process interactive creation from a blank canvas, enabling continuous user participation in the artistic generation process. It maintains precise and controllable content generation through multiple iterations and ensures consistent style stability throughout the oil painting creation process.

\section{Related Works}

\subsection{Stroke Based Rendering with Oil Painting}

Stroke-Based Rendering (SBR) is a graphics technique that enables style transfer while preserving semantic content by placing discrete elements (e.g., brushstrokes) rather than manipulating pixels. 
This approach visualizes the artistic process and supports diverse styles such as pencil sketches~\cite{fu2011animated}, watercolors~\cite{kang2006unified}, and oil paintings~\cite{hu2023stroke, hu2024towards, huang2019learning, liu2021paint, tong2022im2oil}.
Early SBR methods~\cite{hertzmann1998painterly, hertzmann2001paint} often ignore spatial–temporal coherence, producing unnatural stroke sequences. Recent advances using reinforcement learning~\cite{huang2019learning, hu2023stroke} or Transformer~\cite{liu2021paint} improve quality but still lack alignment with authentic human painting workflows. Works such as~\cite{hu2024towards, tong2022im2oil} introduced key principles, e.g., edge-following strokes, coarse-to-fine progression, and semantic region rendering, enhancing realism and supporting style preservation under arbitrary cropping.
Our work builds on this by leveraging state-of-the-art SBR methods to synthesize large-scale oil painting datasets from natural images, addressing the data scarcity problem in oil painting style editing tasks.

\begin{figure*}[t]
\centering
\includegraphics[width=0.95\linewidth]{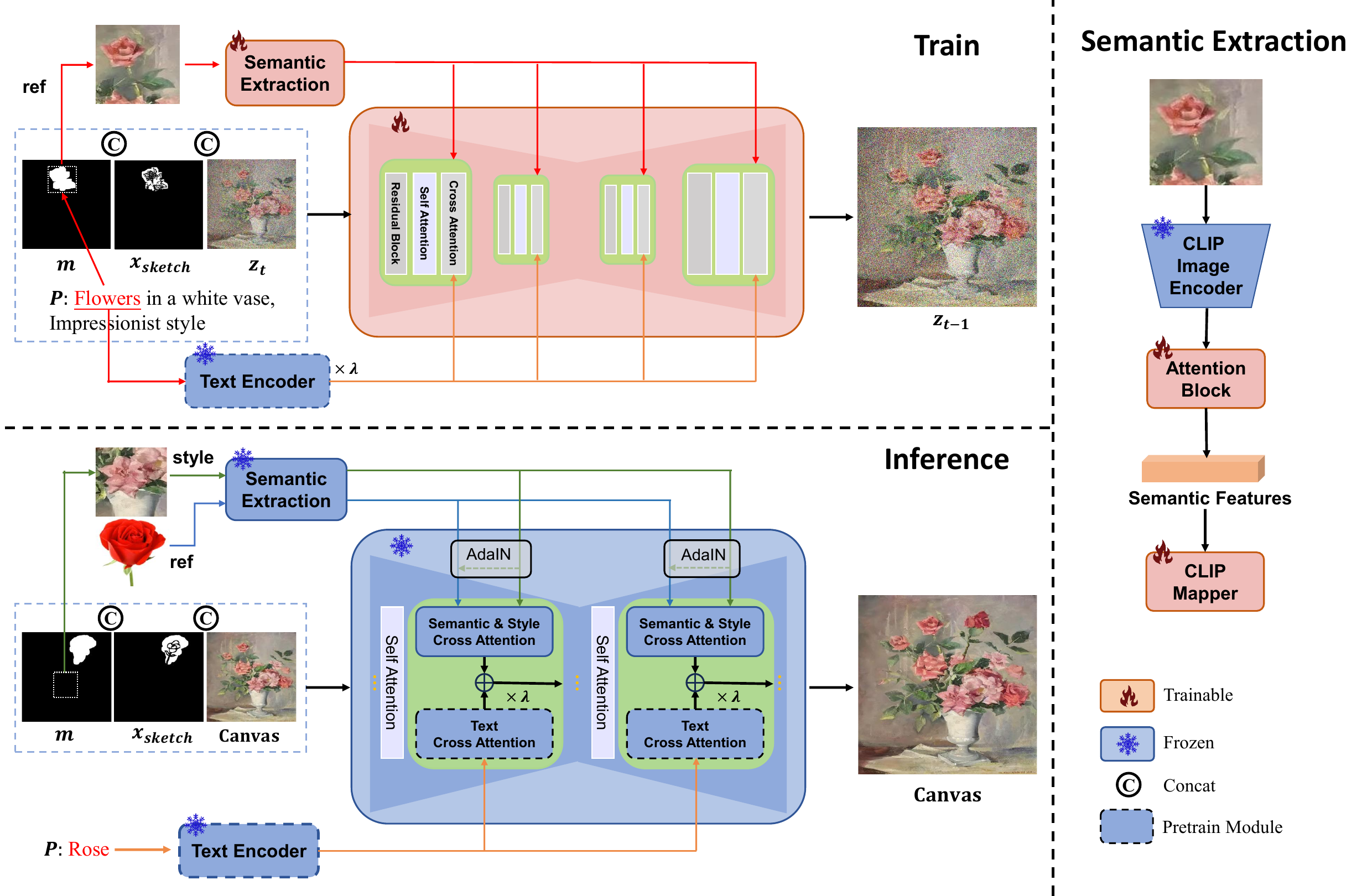}
\caption{Overview of our framework. 
During training, we follow a condition alignment paradigm by feeding the mask and sketch as additional channels into the denoising process. A semantic enhancement strategy is applied to extract fine-grained features as context embedding via cross-attention from the reference image and frozen pretrained text embedding, ensuring detailed semantic fidelity. During inference, we use a training-free AdaIN operator to align source style. Prompt features from the frozen CLIP text encoder are fused with a learnable hyperparameter $\lambda$, enabling visually satisfactory editing.
}
\label{fig:framework}
\end{figure*}

\subsection{Personalized Image Generation and Editing}

Personalized image generation and editing involves tailoring content to user-specific needs. While early GAN-based methods~\cite{goodfellow2020generative} enabled controllable synthesis, they suffered from limited generalization. Diffusion models have since shown superior quality and flexibility. Recent approaches incorporate various conditional inputs—text~\cite{brooks2023instructpix2pix, sheynin2024emu}, masks~\cite{huang2024smartedit, singh2024smartmask, yang2023paint}, layouts~\cite{epstein2023diffusion}, and sketches~\cite{voynov2023sketch, liu2024magicquill}—focusing on text-image alignment~\cite{podell2023sdxl}, identity preservation~\cite{ye2023ip, wu2024infinite}, or style consistency~\cite{wang2023styleadapter, hertz2024style, wang2024instantstyle, liu2024stylecrafter}.
IP-Adapter~\cite{ye2023ip} preserves portrait identity under text guidance but lacks support for fine-grained editing. MagicQuill~\cite{liu2024magicquill} enables multimodal control (text, color, sketch) with spatial flexibility but lacks reference image input for object-level control. Existing methods often rely on coarse, single-modal controls and struggle with conflicts between modalities, hindering joint alignment of semantics, content, and style. To address this, we introduce a multimodal framework with distributed input design and attention control to enable coherent, fine-grained alignment across modalities.

\subsection{Interactive Support for Image Editing}


Interactive support substantially enhances generative models via human-in-the-loop collaboration~\cite{ko2023large}. While existing systems, such as Promptify~\cite{brade2023promptify}, PromptMagician~\cite{feng2023promptmagician} and DesignPrompt~\cite{peng2024designprompt}, pioneer the use of image clustering and attention visualization techniques to concretize human creativity through prompt engineering, their capability remains confined to coarse-grained inpainting. Although MagicQuill~\cite{liu2024magicquill} achieves pixel-level object generation by integrating brushstrokes with a multimodal large language model (MLLM), its reliance on ambiguous textual semantics inherently limits fine-grained intra-element control. The problem faced by recent work is that image editing still essentially relies only on textual modality, and the ambiguous semantics of text leads to biased generation. To resolve this fundamental constraint, our system incorporates reference images and a decoupled training paradigm to harmonize multimodal controls, thereby enabling cross-image fine-grained semantic migration for precision-guided creative expression.

\section{Method}

\subsection{Problem Definition}
To realize a highly available oil painting generation agent, which facilitates a flexible and interactive creative process for oil painting generation and refinement, it is essential to develop a novel, unified framework for oil painting-style image generation and editing. This framework is characterized by two key features: (1) It supports multimodal inputs from users, including reference images, hand-drawn sketches, and text instructions; (2) It enables interactive generation and editing according to user instructions at any stage of the oil painting process (\emph{e.g.} initial blank phase, partially completed stages, or fully finished artwork), and within any specified region of the painting. Thus, we propose to train a diffusion model that generates oil painting style images with the following inputs: a) An initial painting canvas $\mathbf{x_s}\in\mathbb{R}^{H\times W\times 3}$; b) A binary mask $\mathbf{m}\in\{0,1\}^{H\times W}$, which indicates targeted editing regions; c)A reference image $\mathbf{x_r}\in\mathbb{R}^{H^{\prime}\times W^{\prime}\times 3}$ and a regioned sketch $\mathbf{x_{sketch}}\in\mathbb{R}^{H^{\prime}\times W^{\prime}}$ which are responsible for providing structural information of the painted content, where $H^{\prime}$ and $W^{\prime}$ represent the height and width of smallest rectangle to enclose $1$ region in mask $\mathbf{m}$; d) A prompt $\mathbf{P}$ describing the text to draw. After receiving the input above, the model fills the masked regions by incorporating the textual content from the reference image while following the sketch-guided structure and source oil style.

\subsection{Self-supervised Oil Painting Data Preparation}
In order to train the aforementioned generative model, the first step is to construct a multimodal-annotated dataset in the oil painting style. This dataset should include various types of annotations corresponding to the multimodal conditions, wich serves as the foundational resource for enabling the model to understand and generate oil paintings based on different forms of user input.

Specifically, we use DiffusionDB~\cite{wangDiffusionDBLargescalePrompt2022} as our data source and adopt a self-supervised data annotation approach with balanced input conditions to transfer source images to oil painting style images with multimodal annotations. Concretely, given a real image $\mathbf{x_s}$, we obtain its oil painting stylization $\mathbf{x_{oil}}$ through SBR algorithm and its sketch $x_{sketch}$ using edge detection. For a segmented image with foreground object contours, we treat the foreground contours as a binary mask $\mathbf{m}$, and consider the region within the binary mask as the reference image $\mathbf{x_r} = \mathbf{m} \odot \mathbf{x_{oil}}$, along with its corresponding sketch $\mathbf{x_{sketch}} = \mathbf{m} \odot \mathbf{x_{edge}}$. The generated result should naturally exhibit the stylized region within the bounded area. Therefore, our training data consists of pairs $\{(\bar{\mathbf{m}} \odot \mathbf{x_s}, \mathbf{x_{sketch}}, \mathbf{m}, prompt), \mathbf{x_{oil}}\}$, where $\bar{\mathbf{m}} = 1-\mathbf{m}$ represents the complement of $\mathbf{m}$, and $1$ denotes an all-ones matrix with same shape.

Due to limitations in real oil painting datasets, including insufficient quantity and lack of corresponding text descriptions, we adopt the approach proposed by ~\cite{hu2024towards, tong2022im2oil}, utilizing SBR algorithm to stylized real images into oil paintings. A language analysis agent ~\cite{deepseek} is employed to extract subjects among prompts, which are then used as text inputs for Grounded-SAM ~\cite{ren2024grounded} to segment the main objects as mask regions. For sketch images, we employ edge detection model ~\cite{su2021pixel} to extract the primary structures and subsequently binarized these edge structures. To capture richer details, particularly in background regions, we lower the edge detection threshold to reveal finer structural details.

Building upon the accurate foreground object contours obtained through our data preparation pipeline, we apply morphological operations at the foreground object contour edges, expanding outward by 3-6 pixels to form irregularly shaped masks. This random distortions $D_r$ on mask $m$ break the inductive bias and narrow the gap between training and testing scenarios, which indicates $\bar{m}=\textit{1}-D_r(m)$. Previous research ~\cite{yang2023paint} has shown that rectangular masks often fail to accurately represent masked regions in practical applications. 
By incorporating these irregular masks into our training process, our model can generate superior results given masks of various shapes.

We select image-text pairs from DiffusionDB ~\cite{wangDiffusionDBLargescalePrompt2022} and annotate input images and stylize them. 
To ensure the model possesses comprehensive semantic understanding, we adopt a hierarchical data construction strategy: for images with text identifying foreground objects, we construct training data for foreground object completion. For images in which the main subject cannot be extracted from text or the foreground cannot be segmented, we perform random cropping with a cropping ratio of 50\% to construct background completion training data. After filtering, we have formed balanced training data consisting of 40,000 foreground samples and 10,000 background samples.

\subsection{Oil Painting Style Diffusion Model Training}
\textbf{Semantic Preservation and Condition Alignment Paradigm.} To decouple the influence of multimodal conditions (mask, sketch, prompt and reference image) on the generated image, our method separates regional inpainting from semantic preservation conditions at the training stage. 
Compared to other conditions, sketches excel in guiding geometric structures for pixel-level generation control, offering users more refined image editing capabilities. Specifically, in our model architecture, we incorporate masks $m$ and sketches $x_{sketch}$ as input images' new dimensions into the U-Net's denoising process through channel concatenation. It can be expressed as: 
\begin{equation}
    z^{\prime}_t = \mathrm{Concat}(z_t, \mathrm{Resize}(mask), \mathrm{Resize}(x_{sketch}))
\end{equation}
where $\mathrm{Concat}(\cdot, \cdot, \cdot)$, $\mathrm{Resize}$ denote the concatenation and image resize function respectively.

For reference images $x_{ref}$, we propose a novel semantic enhancement strategy: inject encoded image embeddings as context through cross-attention mechanisms while deactivating text cross-attention modules in U-Net in the training stage. This design enables both accurate capture of semantic texture information from reference images and precise control of outline structures through sketches.

Throughout the training phase, we optimize the parameters of the semantic extraction and fine-tune the pre-trained diffusion model. The optimized loss is very similar to the original diffusion loss formulation, with the only difference being the shift from textual conditions to identity conditions that are input as reference image embeddings $c_{ref}$.
\begin{equation}
L_{\mathrm{diffusion}}=\mathbb{E}_{z^{\prime}_{t},t,c_{ref},\epsilon\in N(0,I)}[||\epsilon-\epsilon_{\theta}(z^{\prime}_{t},t,c_{ref})||_{2}^{2}]
\end{equation}

Injecting sketch conditions into the U-Net via channel-wise concatenation demonstrates distinct advantages: it preserves precise spatial structure intact, allows all convolutional layers to directly leverage geometric constraints, and avoids the high computational overhead of attention matrices in cross-attention mechanisms. This conditioning strategy is particularly suitable for our tasks that require strict spatial alignment, such as edge-guided synthesis or layout-controlled generation. In contrast, cross-attention mechanisms are more appropriate for handling semantic features from reference images, as they capture multiscale texture patterns through long-range dependency modeling while maintaining flexibility in spatial semantic relationship.

\noindent\textbf{Semantic Enhancement Training Strategy.} In the feature extraction and fusion stage, we propose a hierarchical feature processing paradigm. Initially, we employ a feature freezing strategy by fixing the parameters of the CLIP image encoder, which extracts patch-level feature representations to form an embedding sequence. Previous research ~\cite{ye2023ip} has demonstrated that features extracted by the CLIP image encoder effectively preserve texture details and structural information from reference images. Subsequently, we introduce a multi-head attention mechanism based on learnable query vectors, which adaptively captures correlations between image patch features at different spatial positions.

Specifically, by designing global query vectors as attention queries, this simple processor can dynamically assign importance weights to different image patches, achieving feature aggregation from local to global perspectives. This attention-based aggregation approach demonstrates significant advantages over direct MLP projection: (1) preservation of spatial relationships between image patches; (2) adaptive adjustment of feature fusion strategies based on input content. Following attention aggregation, we employ a simple MLP network for feature transformation, leveraging non-linear mapping to further enhance feature expressiveness. Finally, the image embeddings are projected into the same dimensionality as text features in the pretrained diffusion.

In summary, the semantic feature embedding $c_{ref}$ extracted by reference image $x_{ref}$ can be expressed as: 
\begin{equation}
\boldsymbol{c}_{\boldsymbol{ref}}=\mathrm{M}_{\mathrm{clip}}\left(\mathrm{E}_{\mathrm{atten}}(\mathrm{E}_{\mathrm{clip}}(\boldsymbol{x}_{ref}))\right)
\end{equation}
where $\mathrm{M}_{\mathrm{clip}}(\cdot)$, $\mathrm{E}_{\mathrm{atten}}(\cdot)$, $\mathrm{E}_{\mathrm{clip}}(\cdot)$ denote the CLIP MLP mapper, multi-head attention extractor, CLIP image encoder.

\noindent\textbf{Modality Fusion.}
For text injection in the original Stable Diffusion, we process through the standard text encoder and use learnable coefficient $\lambda$ to control the mixing ratio between text conditions and image information. Fusion results are passed to subsequent SD blocks.
\begin{equation}
\operatorname{Attn}(Q,K,V)=\operatorname{Attn}(Q,\hat{K}_{ref},\hat{V}_{ref})+\lambda\operatorname{Attn}(Q,K_t,V_t)
\end{equation}
where $K_{t}=\tilde{W}_k^{t} c_{t}$, $V_{t}=\tilde{W}_v^{t} c_{t}$. $c_t$ is the text embedding processed by pre-trained CLIP text encoder. $\tilde{W}_k^{t}$, $\tilde{W}_v^{t}$ correspond to the replication weights of the SD base version.

The attention is computed by the scaled dot-product as:
\begin{equation}
\mathrm{Attn}(Q,K,V) =\mathrm{softmax}\left(\frac{QK^T}{\sqrt{d_k}}\right)V
\end{equation}
where $d_k$ is the dimension of query and key vector.

\subsection{Style Retention Mechanism}
At inference stage, we maintain the training phase methodology: utilizing the trained semantic feature extraction module to capture key semantic and texture information from reference images, while using randomly cropped original images as style conditions. 
Inspired by ~\cite{hertz2024style, wu2024infinite} , we employ an AdaIN operation~\cite{huang2017arbitrary} to further align the style of the synthetic images with the original input images. Specifically, we align the key and value features of the reference image in the cross-attention layer with the key and value features of the input image:
\begin{equation}
\begin{split}
    &\hat{K}_{ref} = \operatorname{AdaIN}(K_{ref}, K_{style})\\
    &\hat{V}_{ref} = \operatorname{AdaIN}(V_{ref}, V_{style})\\
    &K_{ref} =W_k^{ref}c_{ref},\quad K_{style}=W_k^{ref}c_{style} \\
    &V_{ref} =W_v^{ref}c_{ref},\quad V_{style}=W_v^{ref}c_{style}
\end{split}
\end{equation}
where $c_{ref}$ and $c_{style}$ are the corresponding context embeddings of the reference image and style image. The parameters $W_k^{ref}$, $W_v^{ref}$ are the weights of the corresponding fully connected layers.
And the AdaIN operation is defined as:
\begin{equation}
\operatorname{AdaIN}\left(x,y\right)=\sigma\left(y\right)\left(\frac{x-\mu(x)}{\sigma(x)}\right)+\mu(y)
\end{equation}


\section{Experiment}

\begin{figure*}
\centering
\includegraphics[width=0.95\linewidth]{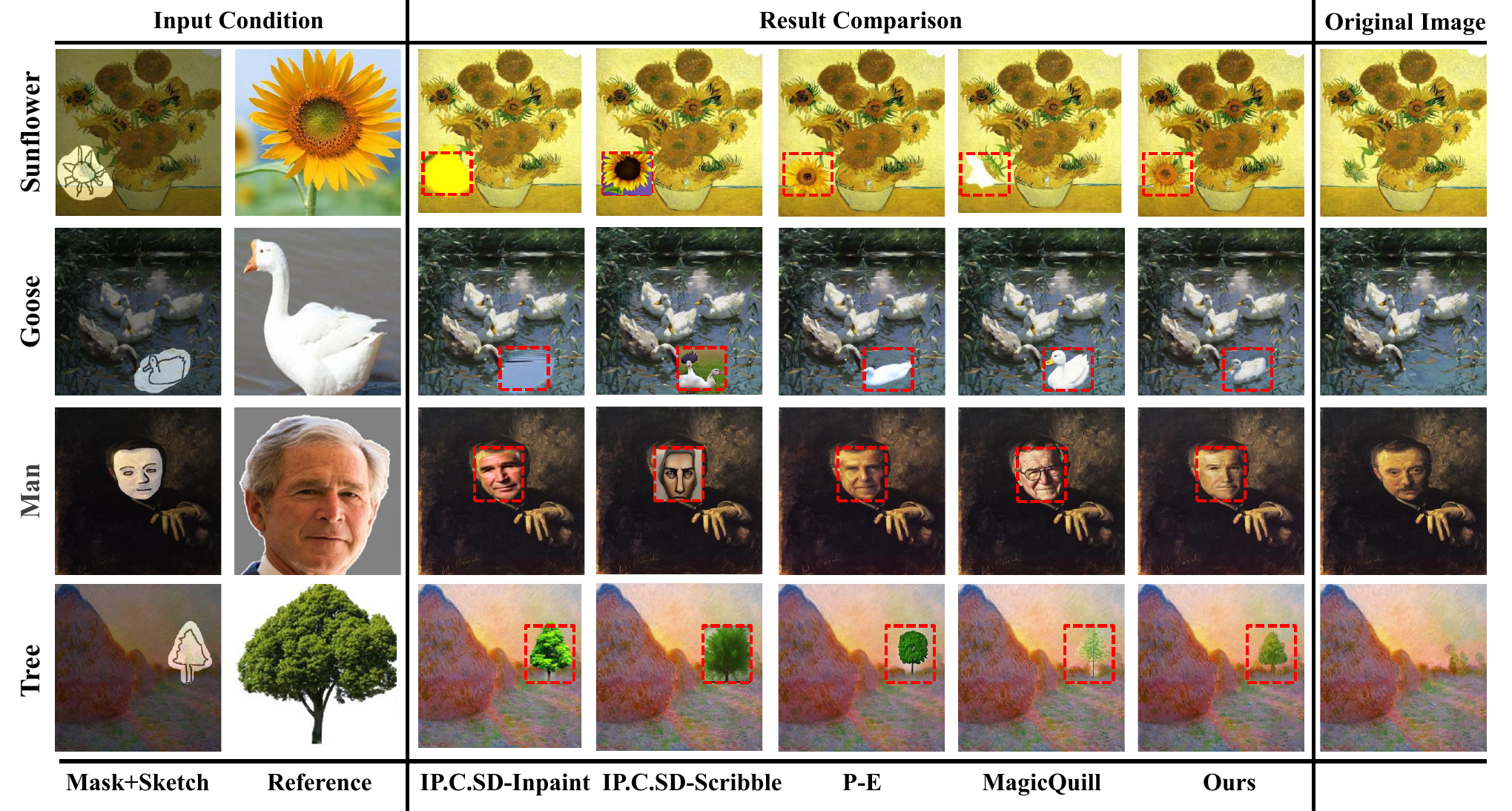}
\caption{Qualitative comparison with state-of-the-art methods. We showcase the inpainting results of previous methods and
ours. As shown, the results of existing methods are plagued with artifacts, irregular edges, and realistic style inconsistencies. Our method achieves significantly better oil paintings than
other methods. Please zoom in to capture more details.}
\label{fig:experiment}
\end{figure*}

\begin{figure*}
\centering
\includegraphics[width=0.9\linewidth]{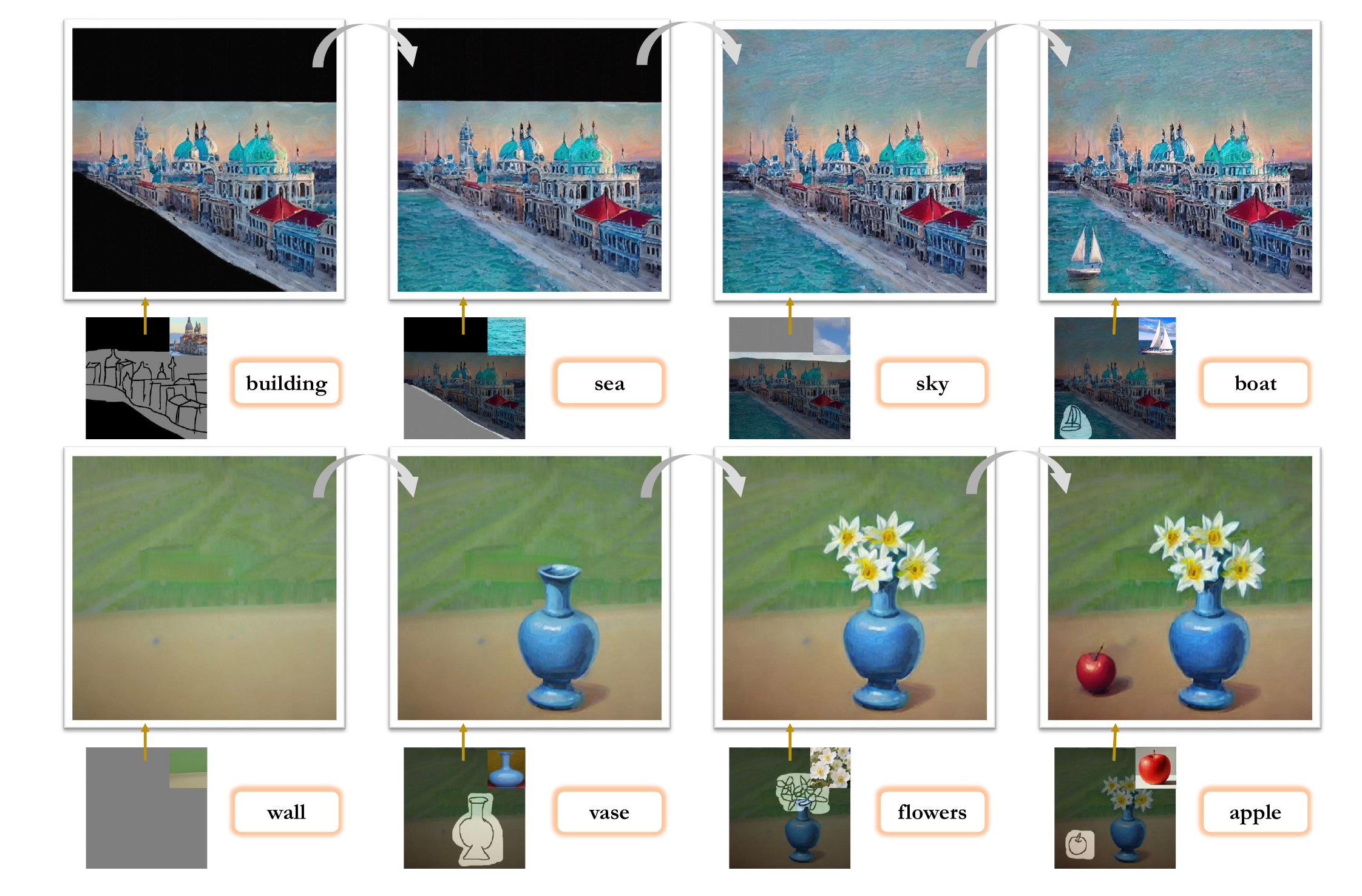}
\caption{Flow Painting Process. We demonstrate how our system generates a high-fidelity oil painting in four steps. }
\label{fig:experiment2}
\end{figure*}

\subsection{Experiment Settings}

\textbf{Training settings.} The model is initialized by pre-trained models ~\cite{rombach2022high}, providing a strong prior guidance at the early stages of training. Input images and reference images are uniformly resized to $512\times512$ and $224\times224$ respectively, and we adopt PiDiNet ~\cite{su2021pixel} as the edge detection model. For the semantic extraction module, we use OpenCLIP ViT-H/14 ~\cite{zhai2019large}, incorporating a trainable global query vector in the attention layer and an 8-head attention mechanism, followed by a CLIP mapper with 768-dimensional output. Training is conducted on 8 NVIDIA 4090 GPUs for 50 epochs. At inference stage, we employ DDIM Sampler ~\cite{song2020denoising} with 50 steps and guidance scale is $3.0$. 

\noindent\textbf{Evaluation.} Our goal is to seamlessly merge the reference image into the source image, ensuring the editing regions are both semantically aligned with the input text and image and blend with the source style. To evaluate these aspects, we use four metrics: (1) CLIP-T ~\cite{radford2021learning} and CLIP-I ~\cite{gal2022image} scores measure the similarity between the editing regions and the reference image; (2) Gram ~\cite{gatys2016image} score assesses feature map style similarity via the Gram matrix; (3) FID ~\cite{heusel2017gans} score, a standard metric for evaluating image quality, using the oil painting dataset from ~\cite{oildataset}; (4) Aesthetic ~\cite{yi2023towards} score, which evaluates the aesthetic quality of the generated images. Methods without corresponding modal omit the scores.

\begin{table*}[t]
\centering
\caption{Quantitative comparison. Our method features the most comprehensive multimodal input, enabling simultaneous alignment of text, image semantics and style. The performance metrics underscore the superiority of our method. Additionally, the quantitative comparisons of ablation study have been shown in the bottom part.}
\label{tab:quantitative}
\begin{tabular}{ccccccccc}
\hline
\multirow{2}{*}{Method} & \multicolumn{3}{c}{Target Alignment} &\multirow{2}{*}{CLIP-T \(\uparrow\)} &\multirow{2}{*}{CLIP-I \(\uparrow\)} & \multirow{2}{*}{Gram \(\uparrow\)} & \multirow{2}{*}{FID \(\downarrow\)} & \multirow{2}{*}{Aesthetic \(\uparrow\)}\\
\cline{2-4}
& Text & Semantics & Style &  &  &  & & \\
\hline
IP-Adapter+SD-ControlNet-Inpaint & \checkmark & \texttimes & \checkmark & 0.862 & - & 0.568 & 268.16 & 5.098\\
\hline
IP-AdapterSD-ControlNet-Scribble& \checkmark & \checkmark & \texttimes & 0.834 & 0.792 & 0.431 & 287.50 & 4.873\\
\hline
Paint-by-Example\cite{yang2023paint} & \texttimes & \checkmark & \texttimes & - & 0.809 & 0.417 & 281.71 & 4.894\\
\hline
MagicQuill\cite{liu2024magicquill} & \checkmark & \texttimes & \checkmark &  0.802 & - & 0.585 & 258.35 & 5.147\\
\hline
Ours & \checkmark & \checkmark & \checkmark & \textbf{0.947}& \textbf{0.909} & \textbf{0.619} & \textbf{234.27} & \textbf{5.322}\\
\hline
\hline
Ours w/o text & \texttimes & \checkmark & \checkmark & - & 0.827 & 0.605 & 260.91 & 5.109\\
\hline
Ours w/o semantic & \checkmark & \texttimes & \checkmark & 0.920 & - & 0.611 & 249.63 & 5.184\\
\hline
Ours w/o style & \checkmark & \checkmark & \texttimes & 0.929 & 0.867 & 0.420 & 279.80 & 5.002\\
\hline
\end{tabular}
\label{tab:comparison}
\end{table*}

\subsection{Comparasion with State-of-the-Art Methods}

We select four representative state-of-the-art diffusion-based image generation and editing methods as baselines, including Paint-by-Example~\cite{yang2023paint}, MagicQuill~\cite{liu2024magicquill} and two IP-Adapter~\cite{ye2023ip}+ControlNet~\cite{zhang2023adding}-based composite multimodal condition models. To guarantee the fairness of the comparison: (1) the original hyper-parameter configurations are strictly maintained; (2) only use official input modalities of each model.


Figure~\ref{fig:experiment} provides a qualitative comparison for editing oil painting images. Existing approaches exhibit conflicts or deficiencies in modality conditions, leading to misalignment with input conditions, inconsistent artistic style preservation, and inadequate handling of complex user requirements. Specifically, IP-Adapter+ControlNet+SD-Scribble introduces visible artifacts at mask boundaries (e.g., "Man") and texture discontinuities (e.g., "Tree"). While SD-Inpaint effectively aligns style and semantics with reference images, it fails to achieve spatial pose control via sketches. Paint-by-Example~\cite{yang2023paint}, lacking multi-modal input capability, is confined to coarse-grained adjustments and often directly transplants objects from reference images (e.g., "Sunflower" and "Goose"), resulting in artistic incoherence. MagicQuill~\cite{liu2024magicquill} enables flexible sketch-based editing and employs MLLM for style alignment but relies on simplistic text prompts that prove insufficient for complex demands. This limitation is evident in the "Sunflower" case, where the method fails to produce meaningful results when confronted with intricate sketch.


In contrast, our approach leverages multimodal conditions to enable flexible and precise editing in the image.
By establishing a three-level intent translation mechanism— "Region Definition(mask)-Layout Planning(sketch)-Precise Semantic Control(reference+text)" — we effectively decouple and leverage multi-modal information to capture the user's intent with precision. Unlike LoRA~\cite{hu2022lora}, which fine-tunes SD model with a limited set of images to align a specific style, our approach emphasizes interactive painting through regional and controllable edits with various styles.

Figure~\ref{fig:experiment2} illustrates our flow painting process, which allows users to easily and effortlessly realize their creative ideas. In the background painting stage, the system generates a semantically appropriate background composition based on text and reference images. When the canvas is blank, the text allows specification of various painting styles. In the foreground painting stage, users can accurately bring their ideas to life on the canvas by combining textual prompts, reference images, and simple sketch, all while maintaining a consistent oil painting style for about 5s inference time. Each step typically takes 2 minutes, while generating a complete image from scratch takes about 8 minutes.

Table~\ref{tab:quantitative} presents a quantitative analysis, highlighting the superiority of our method in generating oil paintings. 
Our method is the first to successfully fuse three modalities of information. It significantly outperforms other model architectures in terms of instruction alignment, style retention and image aesthetic quality.
For more qualitative results, please refer to supplementary material.


\subsection{Ablation Study}

\begin{figure}
\centering
\includegraphics[width=0.95\linewidth]{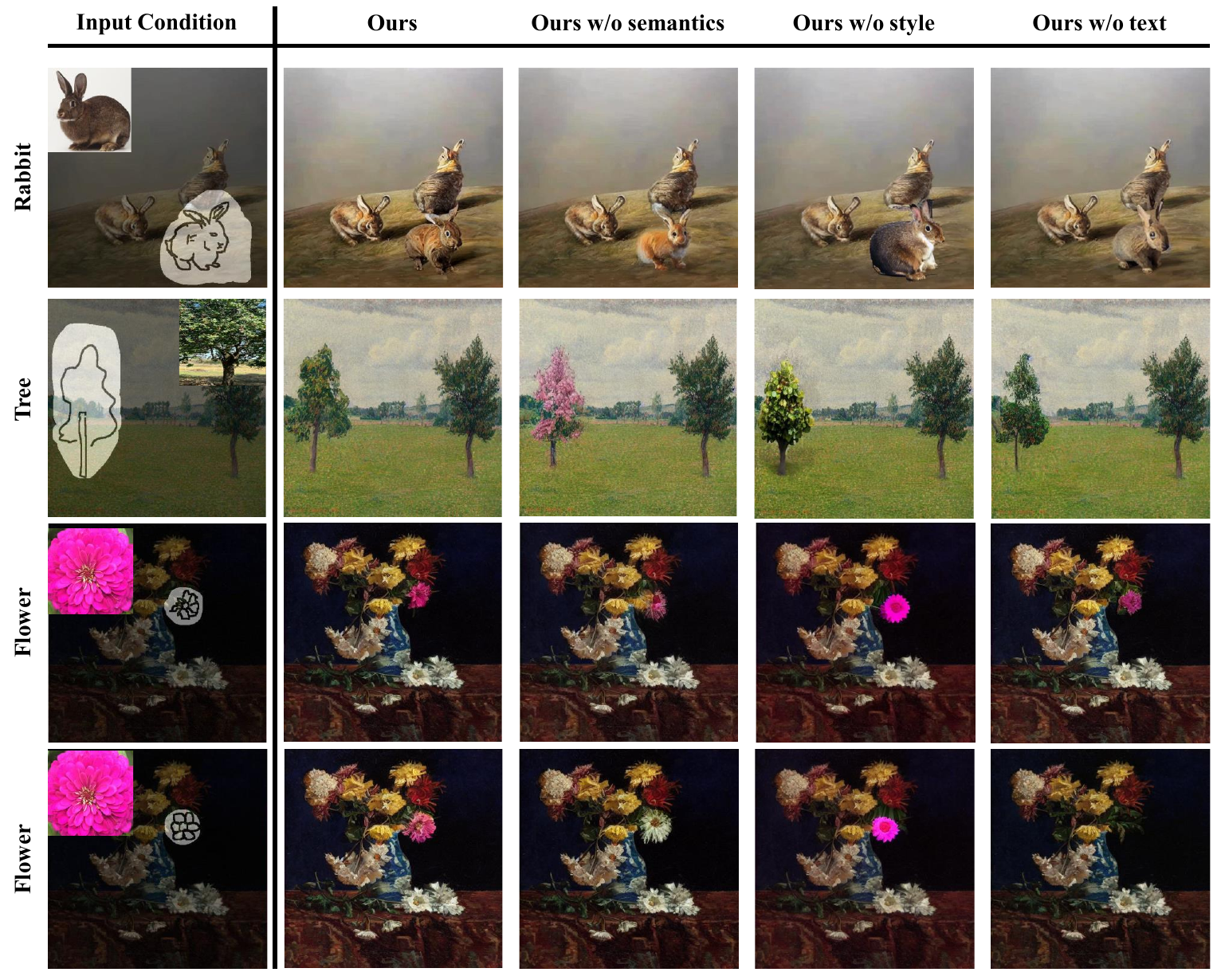}
\caption{Ablation Study. Our method is equipped with a dedicated module for each condition, and the absence of any modality will compromise the generated image fidelity.}
\label{fig:ablation}
\end{figure}

The results of the ablation study, shown in Figure~\ref{fig:ablation}, provide deep insights into the essential roles of each component in our proposed framework for generating high-quality oil painting images.
When the semantic extraction module is removed, the generated image will rely solely on the text to provide texture and color information. This leads to a loss of fine-grained control over the image's regions, particularly in areas where precise details are crucial, and results in a more generic and less structured output. Without the style operator, the generated images adopt a more realistic style, moving away from the intended oil painting aesthetic. 
Without text input, the critical function of providing high-level semantic guidance will be absent. The text ensures robust output generation when spatial and visual information conflict, by bridging deep connections between the image (reference) and layout (sketch). In the absence of text modality, most generated images lack fine-grained texture details, as shown in the third row. The fourth row presents failure case: ambiguous input sketches cause high ambiguity, leading to modal conflict with the reference images and resulting in image generation failure. This highlights the importance of text. The qualitative comparison shown in Table~\ref{tab:quantitative} demonstrates the effectiveness of our modules. 

\section{Conclusion}

In conclusion, we have presented a novel framework for interactive oil painting generation and editing that leverages multimodal conditions. By customizing text, sketch, and reference image conditions, we provide users with fine-grained control over the creation and modification of oil paintings. Our framework integrates a condition alignment paradigm during training, along with a semantic enhancement strategy, to preserve the intricate details of the reference image. The style retention mechanism during inference ensures the generated content aligns with the desired oil painting style and user instructions.
We introduce a self-supervised pipeline for generating an unlimited oil painting dataset from the real dataset. Extensive experiments validate the effectiveness of our approach, demonstrating its ability to generate oil paintings with high fidelity, accurate instruction alignment, and robust style preservation. Ultimately, our method offers a comprehensive solution for interactive oil painting generation, enabling users to create artwork from scratch or edit existing paintings with precision and artistic control.

\bibliography{aaai2026}

\clearpage



\appendix

\section{Supplementary Material}
\subsection{Preliminaries}

\textbf{Latent Diffusion Model.}
Latent Diffusion Models (LDM) first employs an autoencoder consisting of an encoder $\mathcal{E}$ and a decoder $\mathcal{D}$ to compress images into lower-dimensional latent representations before training a diffusion model, dramatically reducing computational requirements while maintaining high-quality generation capabilities. Then a denoising U-Net architecture is trained to denoise a noise $\epsilon$ following normal distribution into realistic latent representation $z=\mathcal{E}(x)$ through $t$ timesteps.

From a variational perspective, the LDM objective function can be simplified as:
\begin{equation}
\label{eq:Loss}
\mathbf{L}=\mathbb{E}_{\mathbf{z}_t,c,\epsilon,t}(||\epsilon-\epsilon_\theta(\mathbf{z}_t,c,t)||_2^2)
\end{equation}
where $\epsilon_\theta$ means the function of the denoising U-Net and $c$ denotes the embeddings of conditional information.

In conditional diffusion models, classifier guidance is a technique based on the gradients of pre-trained classifiers that effectively balances fidelity and diversity in generated images. To simplify this process, researchers proposed classifier-free guidance, which jointly trains conditional and unconditional diffusion models by randomly dropping condition $c$ during training. During the sampling stage, the model combines predictions from both the conditional model $\epsilon_\theta(x_t,c,t)$ and the unconditional model $\epsilon_\theta(x_t,t)$ to calculate the final noise prediction:
\begin{equation}
\hat{\epsilon}_\theta(\boldsymbol{x}_t,\boldsymbol{c},t)=w\boldsymbol{\epsilon}_\theta(\boldsymbol{x}_t,\boldsymbol{c},t)+(1-w)\boldsymbol{\epsilon}_\theta(\boldsymbol{x}_t,t)
\end{equation}
The scalar parameter $w$ (commonly known as guidance scale or guidance weight) is used to adjust how closely the generated results match condition $\boldsymbol{c}$, it further enhances conditional generation control, ensuring generated results better conform to user-specified conditions while maintaining reasonable diversity.

\noindent\textbf{CLIP Vision Model.} The vision model in CLIP is widely used for extracting features from visual data in Text-to-Image (T2I) models. To process a image, such as our reference image $I_r \in \mathbb{R}^{H \times W \times C}$, where $H$, $W$, and $C$ denote the height, width, and channels, respectively, the model first flattens the image into a sequence of patches $I_p \in \mathbb{R}^{N \times (P^2 \times C)}$. In this case, $P$ refers to the patch size, and $N$, which equals $HW/P^2$, represents the sequence length. The model then feeds the sequence of patches into a vision embedding module, which computes embeddings using a linear projection matrix $E \in \mathbb{R}^{P^2 \times C \times D}$. Additionally, the model appends a class embedding $E_{cls} \in \mathbb{R}^{1 \times D}$ to the vision embeddings before combining them with a position embedding $E_{pos} \in \mathbb{R}^{(N+1) \times D}$. The embedding process can be expressed as:
\begin{equation}
E_{I_r} = [E_{cls}, I_r^0 E, I_r^1 E, \cdots,I^{N-1}_r E] + E_{pos}.
\end{equation}
In the Semantic Enhancement Training Strategy, we propose a hierarchical feature processing paradigm. Specifically, we first extract patch-level feature representations from the penultimate layer of the CLIP vision encoder, which retain rich local semantic information. To further capture the relationships among patches across different spatial locations, we use a multi-head attention mechanism based on learnable query vectors. In particular, a global query vector is designed to serve as the attention query, allowing the model to dynamically assign importance weights to different patches and thereby aggregate features from local to global.

Compared to simple average pooling, this attention-based aggregation approach offers significant advantages: it not only preserves the spatial relationships between patches, but also adaptively adjusts the feature fusion strategy based on the input content. Following attention aggregation, we apply a two-layer MLP network as CLIP Mapper for feature transformation, enhancing the expressiveness of the features through nonlinear mapping and normalization operations. This design remains end-to-end trainable and effectively mitigates the information loss commonly seen in traditional average pooling methods, providing more discriminative feature representations for downstream tasks.


\subsection{Model Architecture Details}

\textbf{Dataset Preparation.}
DiffusionDB ~\cite{wangDiffusionDBLargescalePrompt2022} is a large-scale text-to-image prompt dataset, containing 14 million images generated by Stable Diffusion based on prompts and hyperparameters specified by real users. Its strength lies in the massive scale of text-image pairs, which allows for accurate identification of key painting areas in generated images by extracting relevant keywords from the text. However, a drawback of DiffusionDB is that these generated images lack the texture and distinctive features typical of oil paintings. Therefore, an SBR algorithm ~\cite{tong2022im2oil} is required to perform style transfer on regular images while preserving all semantic content. The algorithm used for annotation is presented in Alg~\ref{alg:dataset}.

\begin{algorithm}[!htbp]
\caption{Oil Painting Dataset Preparation}
\label{alg:dataset}
\textbf{Input}: DiffusionDB (Input Dataset)\\
\textbf{Output}: Oil painting style train dataset $\mathcal{D}$

\begin{algorithmic}[1]
\STATE Initialize dataset: $\mathcal{D} = []$
\FOR{$i = 1$ \textbf{to} $N$}
    \STATE Get image and text from dataset: $\boldsymbol{x_s}$, $prompt$ = DiffusionDB[$i$]
    \STATE Stylize input image: $\boldsymbol{x_{oil}} = Style(\boldsymbol{x_s})$
    \STATE Edge detection: $\boldsymbol{x_{sketch}} = Edge(\boldsymbol{x_s})$
    \STATE Extract subjects: $\boldsymbol{P_{obj}} = LLM(prompt)$
    \STATE Segment input image with object prompt: $\{\boldsymbol{m_1}, ..., \boldsymbol{m_n}\} = GSAM(\boldsymbol{x_s}, \boldsymbol{P_{obj}})$
    \FOR{each object segment $\boldsymbol{m_j}$ from ${m_1}$ to ${m_n}$}
        \STATE Apply morphological operation: $\boldsymbol{m} = \boldsymbol{D_r}(\boldsymbol{m^j})$
        \STATE Compute the complement: $\boldsymbol{\bar{m}} = 1 - \boldsymbol{m^j}$
        \STATE Update sketch: $\boldsymbol{x_{sketch}} = \boldsymbol{m} \odot \boldsymbol{x_{sketch}}$
        \STATE $\mathcal{D} \gets \mathcal{D} \cup \{(\boldsymbol{\bar{m}} \odot \boldsymbol{x_s}, \boldsymbol{x_{sketch}}, \boldsymbol{m}, prompt), \boldsymbol{x_{oil}}\}$
    \ENDFOR
\ENDFOR
\STATE \textbf{return} Oil painting style train dataset $\mathcal{D}$
\end{algorithmic}
\end{algorithm}

\begin{figure}
\centering
\includegraphics[width=\linewidth]{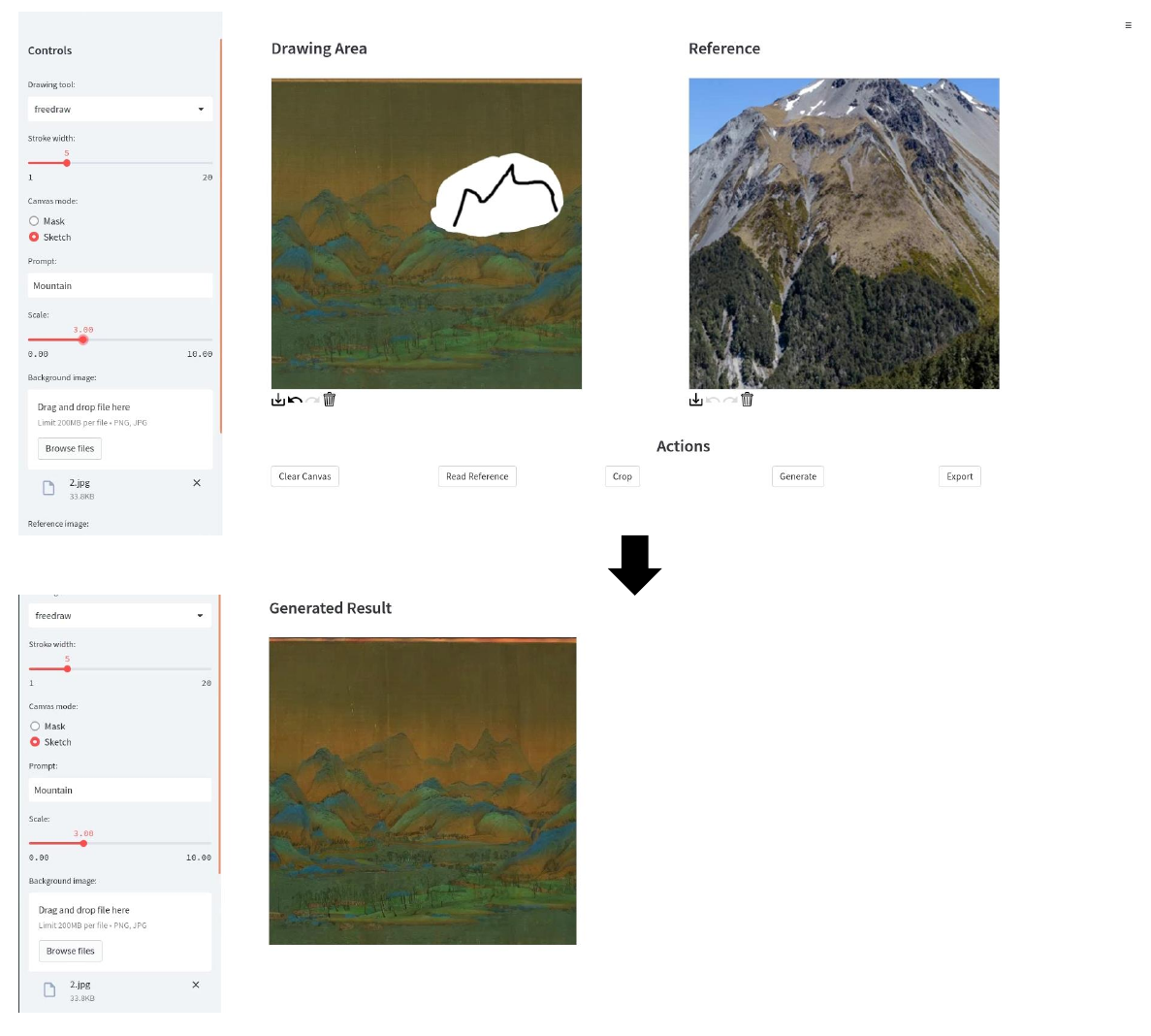}
\caption{Interactive Painting Editing/Generation Interface.}
\label{fig:interactive}
\end{figure}

\noindent\textbf{Details in Interactive Editing.}
We built an interactive painting editing and generation interface based on Streamlit to implement Flow Painting, as shown in Fig~\ref{fig:interactive}. When editing an existing image, users initiate the process by uploading both the source and reference images. They then switch to Canvas mode to separately define masks and sketches. Following this, a prompt is entered in the sidebar, and clicking "Generate" produces the edited result below. For creation from scratch, the source image upload is omitted while all subsequent operations—switching to Canvas mode, defining masks and sketches, entering prompts, and generating the image—remain consistent. To achieve stylistic diversity, users may incorporate relevant style descriptors into their prompts during the first generation stage. The interactive editing algorithm process is illustrated in Alg~\ref{alg:interactive_image_editing_system}. To finalize, a concise semantic prompt is entered in the command input field to configure generation logic, enabling iterative creative outputs within approximately 2 minutes. 

\begin{algorithm}[!htp]
\caption{Interactive Image Editing System}
\label{alg:interactive_image_editing_system}
\textbf{Input}: Canvas $\boldsymbol{C}$, Mask $\boldsymbol{M}$, Sketch $\boldsymbol{S}$, Reference $\boldsymbol{R}$, Prompt $P$\\
\textbf{Output}: Final canvas $\boldsymbol{C_{curr}}$
\begin{algorithmic}[1]
\STATE Initialize previous canvas: $\boldsymbol{C_{prev}} = InitCanvas()$
\STATE Initialize current canvas: $\boldsymbol{C_{curr}} = \boldsymbol{C_{prev}}$
\WHILE{True}
    \STATE Get user input: $\boldsymbol{I_{user}} = GetUserInput()$
    \IF{$\boldsymbol{I_{user}}$ == "break"}
        \STATE \textbf{break}
    \ENDIF
    \STATE Extract current mask: $\boldsymbol{M_{cur}} = \boldsymbol{I_{user}}[mask]$
    \STATE Extract current sketch: $\boldsymbol{S_{cur}} = \boldsymbol{I_{user}}[sketch]$
    \STATE Extract current reference: $\boldsymbol{R_{cur}} = \boldsymbol{I_{user}}[ref]$
    \STATE Extract current prompt: $P_{cur} = \boldsymbol{I_{user}}[prompt]$
    \STATE $\boldsymbol{C_{temp}} = Inference(\boldsymbol{M_{cur}}, \boldsymbol{S_{cur}}, \boldsymbol{R_{cur}}, P_{cur})$
    \STATE Display result to user: $Display(\boldsymbol{C_{temp}})$
    \STATE Get user confirmation: $conf = GetConfirm()$
    \IF{$conf$ == True}
        \STATE Update previous canvas: $\boldsymbol{C_{prev}} = \boldsymbol{C_{curr}}$
        \STATE Update current canvas: $\boldsymbol{C_{curr}} = \boldsymbol{C_{temp}}$
    \ELSE
        \STATE Revert to current canvas: $\boldsymbol{C_{temp}} = \boldsymbol{C_{curr}}$
    \ENDIF
\ENDWHILE
\STATE \textbf{return} Final canvas $\boldsymbol{C_{curr}}$
\end{algorithmic}
\end{algorithm}

\subsection{Experiment Results}

\noindent\textbf{Experiment Settings.} Training is conducted on a system equipped with an Intel Xeon Gold 6150 @ 2.70GHz CPU, running Ubuntu 24.04.2 LTS, and 8 NVIDIA RTX 4090 GPUs. During training, each GPU uses approximately 24.3 GB of memory. We selected 50,000 training images and 3,000 test images from DiffusionDB, maintaining the dataset in paired format as $\{(\bar{\mathbf{m}} \odot \mathbf{x_s}, \mathbf{x_{sketch}}, \mathbf{m}, \text{prompt}), \mathbf{x_{oil}}\}$.

For testing, we introduce an interactive evaluation mechanism that provides qualitative insights through comparative analysis of the completion of the user creative intent. Specifically, for oil painting images in diverse artistic styles, users perform semantic editing on arbitrary image regions by supplying sketches, reference images, and text prompts. 

Among the four baseline models we compare, Paint-by-Example ~\cite{yang2023paint} and MagicQuill ~\cite{liu2024magicquill} are evaluated using their official recommended settings. The other two models are integrated versions of IP-Adapter ~\cite{ye2023ip} and ControlNet ~\cite{zhang2023adding}, with the primary difference lying in the type of condition input used by ControlNet. Specifically, ControlNet-Inpaint and ControlNet-Scribble are both based on Stable Diffusion 1.5, as determined by the pretrained weights provided by their official sources.

\noindent\textbf{User Study.}
To further demonstrate the superiority of our method, we randomly select 50 generated samples and recruit 35 volunteers to rank the editing and generation quality of both state-of-the-art methods and ours across three aspects: text alignment, style consistency, and visual quality. As shown in Table~\ref{tab:user}, the edited oil paintings produced by PaintFlow are consistently preferred in all three aspects.

\begin{table}[ht]
\caption{User study results summarized from rankings of
35 volunteers. Our model ranks first in all three metric.}
\centering
\label{tab:user}
\begin{tabular}{cccc}
\hline
\textbf{Methods} & \textbf{Semantic}$\downarrow$ & \textbf{Style} $\downarrow$ & \textbf{Quality} $\downarrow$ \\
\hline
IPA.+SDC.-Inpaint & 2.77 & 2.80 & 2.86\\
\hline
IPA.+SDC.-Scribble & 4.03 & 3.61 & 3.79\\
\hline
Paint-by-Example & 2.63 & 4.51 & 4.26\\
\hline
MagicQuill & 4.01 & 2.43 & 2.54\\
\hline
Ours & \textbf{1.57} & \textbf{1.66} & \textbf{1.54} \\
\hline
\end{tabular}
\end{table}

\begin{figure*}
\centering
\includegraphics[width=\linewidth]{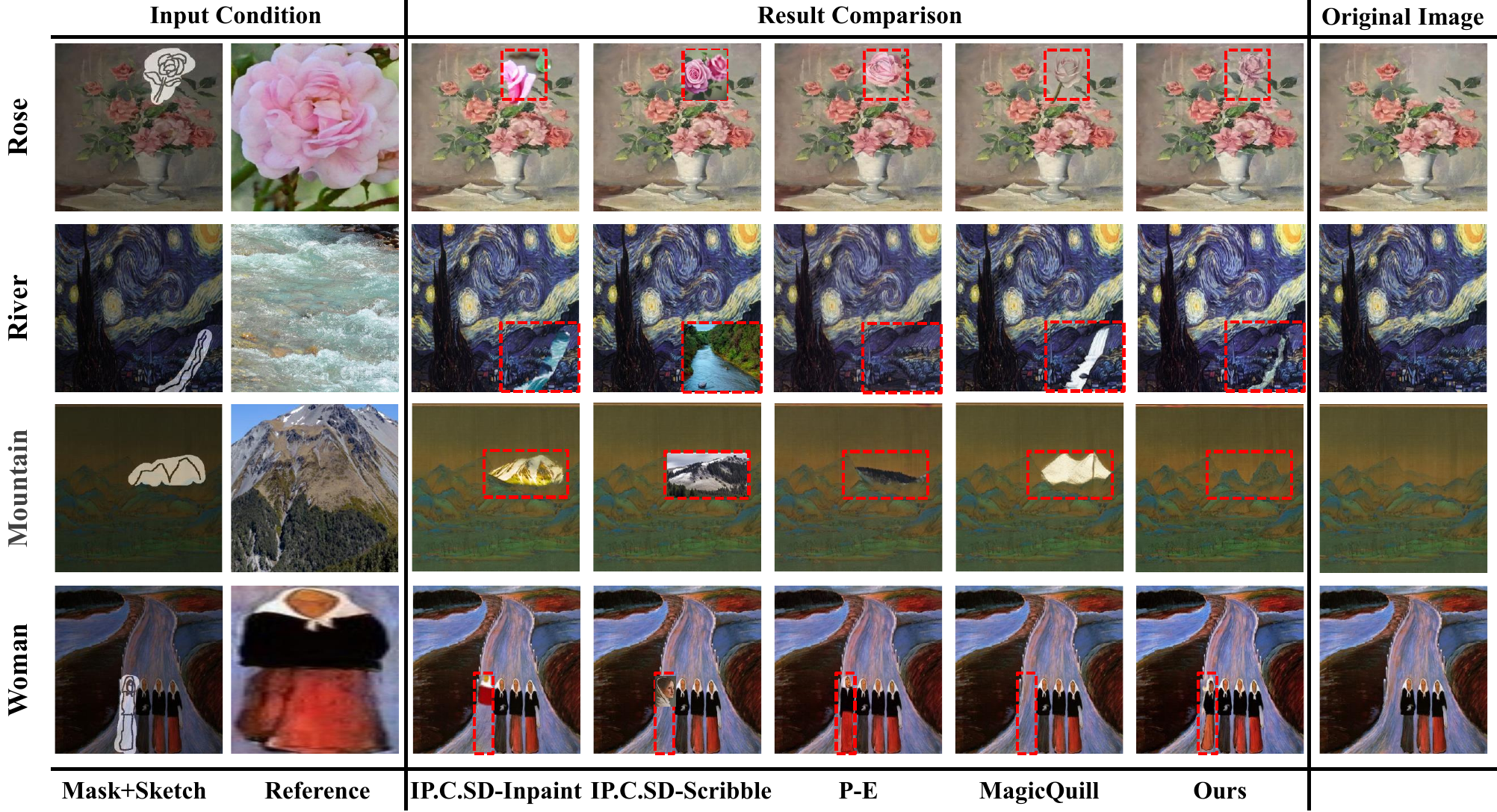}
\caption{More qualitative comparison with state-of-the-art methods.}
\label{fig:supple_experiment}
\end{figure*}

\begin{figure*}
\centering
\includegraphics[width=0.95\linewidth]{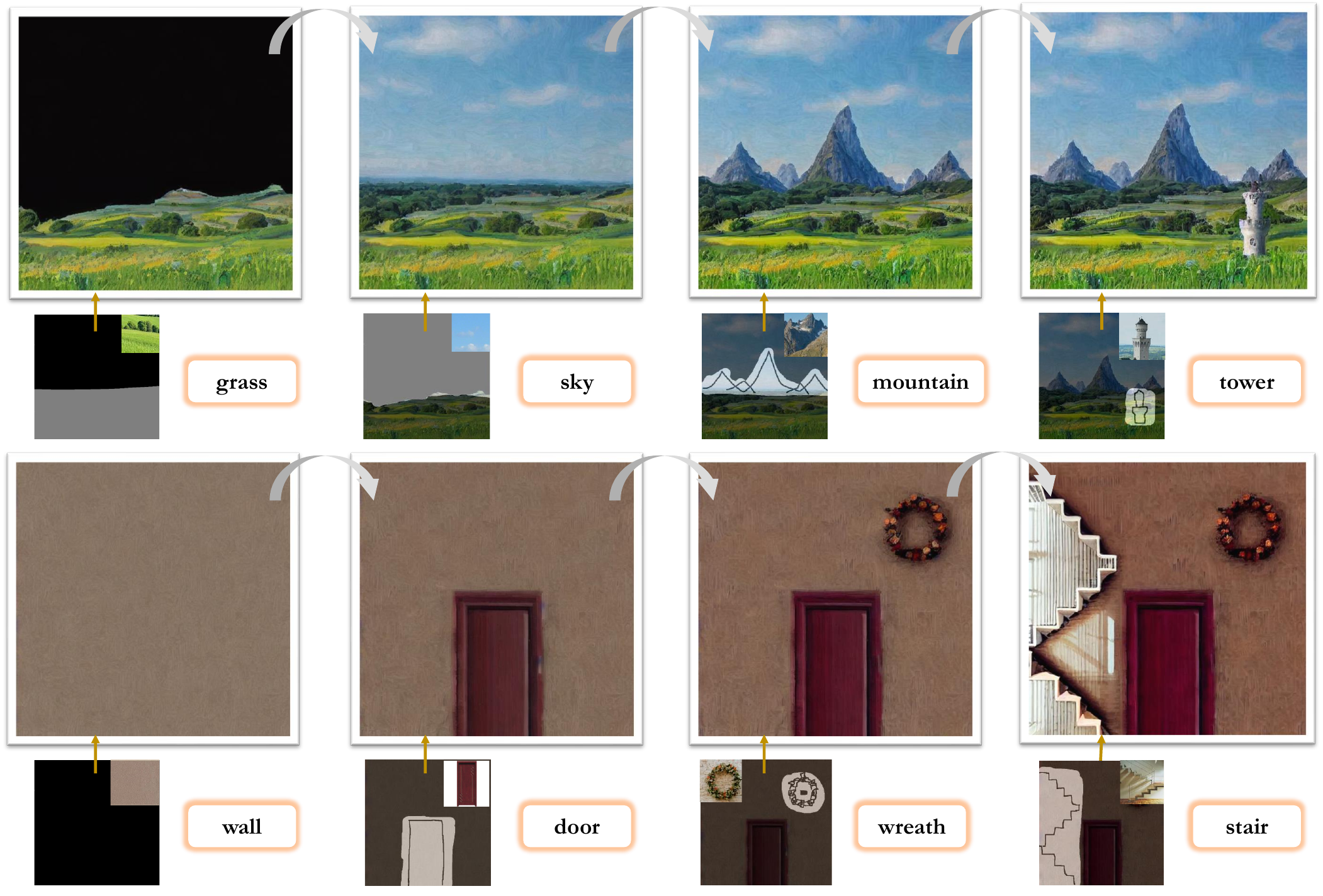}
\caption{More Oil Flow Painting Process. }
\label{fig:supple_experiment2}
\end{figure*}

\begin{figure*}[!htp]
\centering
\includegraphics[width=\linewidth]{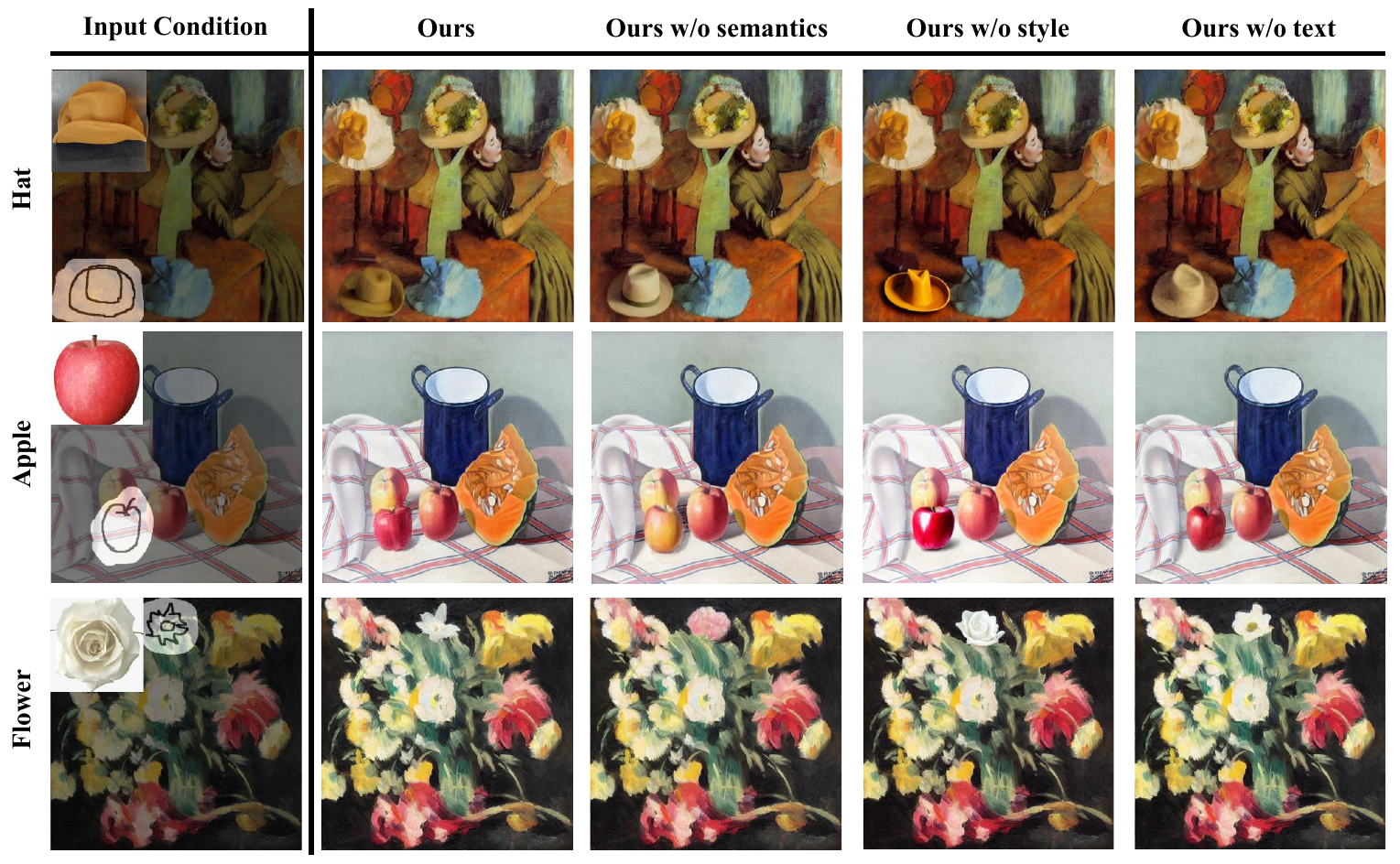}
\caption{Ablation study results.}
\label{fig:supple_ablation}
\end{figure*}

\begin{figure*}
\centering
\includegraphics[width=0.95\linewidth]{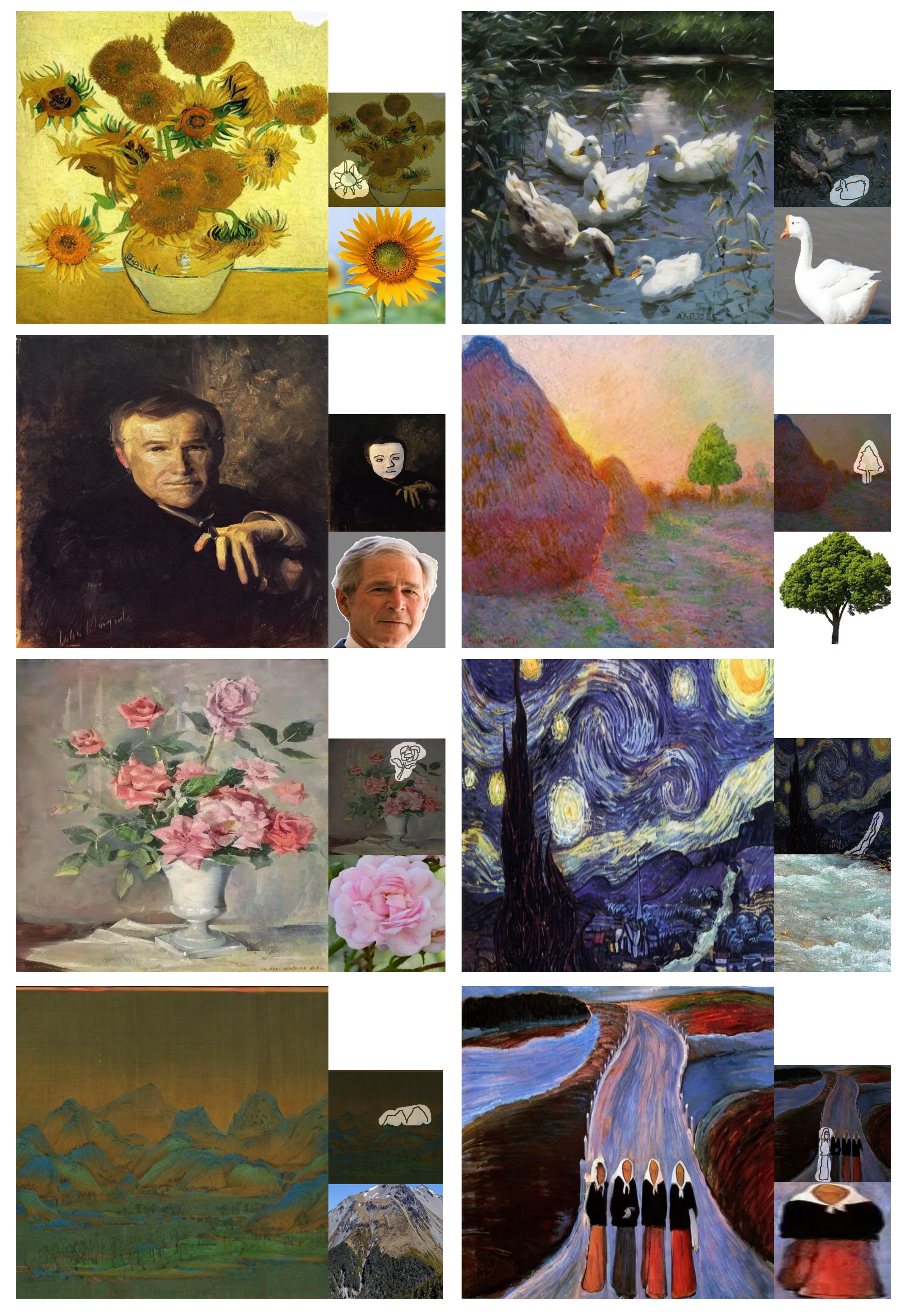}
\caption{Oil paintings editing through our method.}
\label{fig:supple_pic}
\end{figure*}

\noindent\textbf{More Qualitative Comparison Results.}
We provide extensive qualitative comparisons with state-of-the-art methods in Fig~\ref{fig:supple_experiment} and Fig~\ref{fig:supple_experiment2}, alongside further ablation study results in Fig~\ref{fig:supple_ablation}. Finally, we demonstrate high-resolution paintings generated by our method in Fig~\ref{fig:supple_pic}.

\end{document}